\documentclass[10pt,twocolumn,letterpaper]{article}

\usepackage{cvpr}              

\usepackage[accsupp]{axessibility} 

\usepackage{epsfig}
\usepackage{graphicx}
\usepackage{amsmath}
\usepackage{amssymb}
\usepackage{multirow}

\usepackage{amsfonts}
\usepackage{mathrsfs}
\usepackage{booktabs}

\usepackage{paralist} 

\usepackage[dvipsnames]{xcolor}

\definecolor{cvprblue}{rgb}{0.21,0.49,0.74}
\usepackage[pagebackref,breaklinks,colorlinks,citecolor=cvprblue]{hyperref}

\def\paperID{9714} 
\def\confName{CVPR}
\def\confYear{2024}

\title{Flatten Long-Range Loss Landscapes for Cross-Domain Few-Shot Learning}

\author{Yixiong Zou, Yicong Liu, Yiman Hu, Yuhua Li, Ruixuan Li\thanks{Corresponding author. Code is at https://github.com/Zoilsen/FLoR.}\\
	School of Computer Science and Technology, Huazhong University of Science and Technology\\
	{\tt\small \{yixiongz, smnight, m202273659, idcliyuhua, rxli\}@hust.edu.cn}
}

\begin{document}
	\maketitle
	\begin{abstract}
		
		
		Cross-domain few-shot learning (CDFSL) aims to acquire knowledge from limited training data in the target domain by leveraging prior knowledge transferred from source domains with abundant training samples. CDFSL faces challenges in transferring knowledge across dissimilar domains and fine-tuning models with limited training data.
		To address these challenges, we initially extend the analysis of loss landscapes from the parameter space to the representation space, which allows us to simultaneously interpret the transferring and fine-tuning difficulties of CDFSL models.
		We observe that sharp minima in the loss landscapes of the representation space result in representations that are hard to transfer and fine-tune. Moreover, existing flatness-based methods have limited generalization ability due to their short-range flatness.
		To enhance the transferability and facilitate fine-tuning, we introduce a simple yet effective approach to achieve long-range flattening of the minima in the loss landscape.
		This approach considers representations that are differently normalized as minima in the loss landscape and flattens the high-loss region in the middle by randomly sampling interpolated representations. We implement this method as a new normalization layer that replaces the original one in both CNNs and ViTs. This layer is simple and lightweight, introducing only a minimal number of additional parameters. Experimental results on 8 datasets demonstrate that our approach outperforms state-of-the-art methods in terms of average accuracy. Moreover, our method achieves performance improvements of up to 9\% compared to the current best approaches on individual datasets.
		Our code will be released.
		
	\end{abstract}
	
	\vspace{-0.1cm}
	\section{Introduction}

	Cross-domain few-shot learning (CDFSL) is introduced as a solution to mitigate the need for extensive training data on the target domain. It leverages knowledge transferred from non-overlapping source domain datasets that have sufficient training data~\cite{DBLP:journals/corr/abs-1904-04232,guobroader}. 
	It poses two primary challenges due to domain shifts between the source and target domains. (1) \textbf{Transferring}: overcoming domain shifts to transfer knowledge from source to target datasets. (2) \textbf{Fine-tuning}: leveraging the transferred knowledge to learn from limited training data in the target domain.
	Numerous approaches~\cite{Zhou_2023_CVPR,li2022ranking} have been proposed to address these challenges.
	However, these challenges still remain unsolved.
	
	\begin{figure}[t]
		\centering
		\includegraphics[width=1.0\columnwidth]{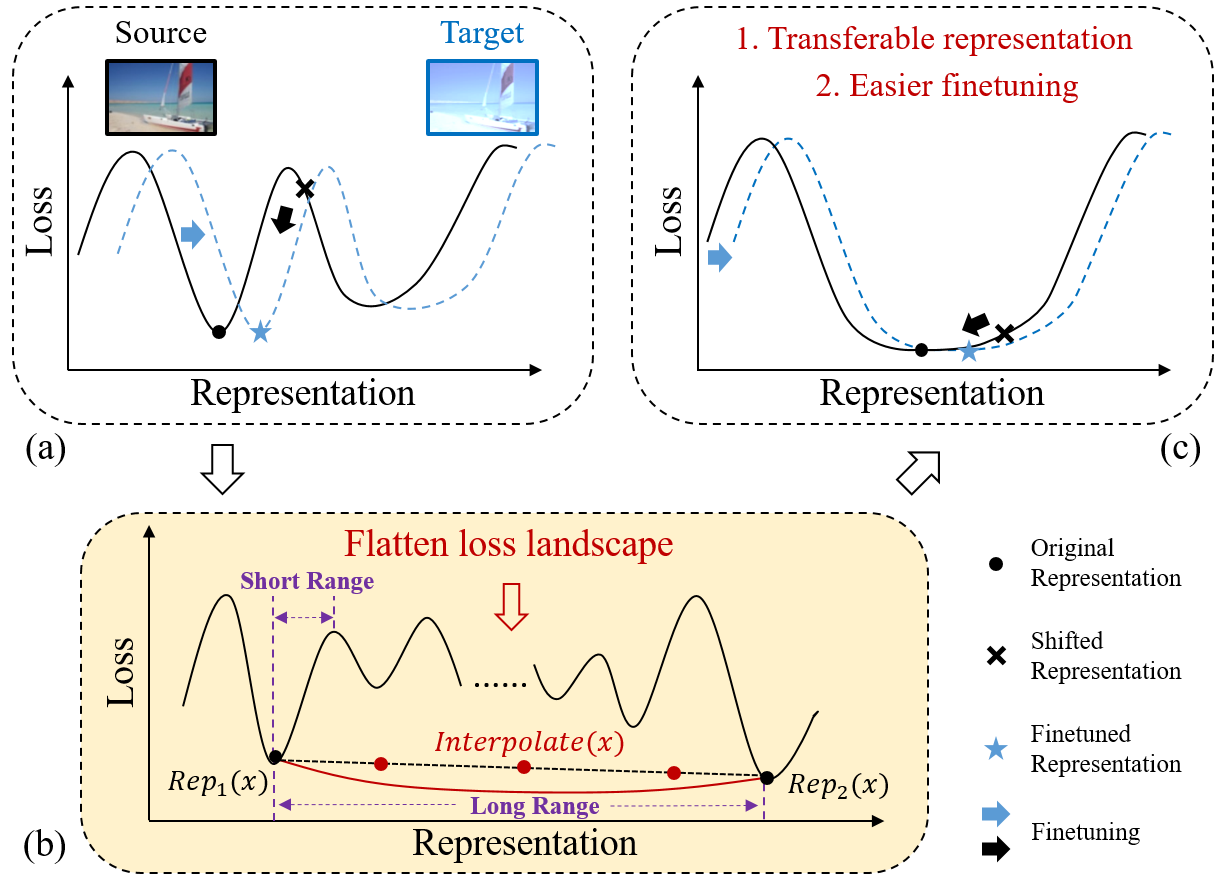}\vspace{-0.2cm}
		\caption{
			(a) 
			Representation-space loss landscape (RSLL): Given an input sample, the model maps it into a representation space, where effective representations correspond to low classification losses, i.e., minima in the landscape.
			Since domain shifts can be reflected by the landscape shift and representation shift, a sharp minimum in the landscape corresponds to a representation vulnerable to domain shifts, making the training and finetuning difficult. 
			(b) We can easily find different minima in the RSLL, which covers a longer range than current flatness-based methods. This inspires us to flatten a long-range loss landscape by randomly interpolating these minima. 
			(c) Given a flattened minimum, the representation and the model are more transferable against domain shifts and easier to be finetuned on the target domain.}\vspace{-0.3cm}
		\label{fig:motivation}
	\end{figure}
	
	Currently, several works~\cite{foret2021sharpnessaware, keskar2017largebatch, zhang2023gradient} have attempted to address the challenge posed by domain shifts by investigating the flatness of the loss landscape. However, these works primarily concentrate on the parameter space, which can not reflect the presence of domain shifts directly. To tackle the CDFSL problem, we extend this concept from the parameter space to the representation space, which provides a more direct means of capturing domain shifts. 
	
	In the source domain training phase, a model is trained to extract effective representations for each input training sample. The representation for a source-domain sample can be seen as a point in the representation space, where good representations correspond to low classification losses. Consequently, we can draw landscapes to map each representation point to its classification loss, with effective representations corresponding to minima, as illustrated in Fig.~\ref{fig:motivation}a.
	Note that this sample can also be represented by its original pixels, which enables the direct analysis of domain shifts.
	
	Since the model is trained on the source domain dataset, the representation extracted by this model is effective enough for recognition. Therefore, each source-domain representation (in feature space or pixel space) extracted by this model corresponds to a minimum in the loss landscape.
	By applying domain shifts to this source-domain sample, its representation may shift from the original minimum (black dot in Fig.~\ref{fig:motivation}) to another point (black cross in Fig.~\ref{fig:motivation}). Compare Fig.~\ref{fig:motivation}a and Fig.~\ref{fig:motivation}c, intuitively, we can see 
	(1) a flat minimum would tolerate larger representation shifts, making the transferring easier; and (2) a flat minimum would reduce the high-loss region between the source and target representations, making the finetuning easier.
	
	Therefore, our goal is to flatten the loss landscape in the representation space during source-domain training to enhance both transferring and fine-tuning. However, we find that the current flatness-based method~\cite{foret2021sharpnessaware} has limited performance on the CDFSL task. By experiments, we conclude this is because current methods rely on the flatness of the landscape in the vicinity of the minimum, which covers only a short range. Therefore, it struggles to effectively address large domain shifts in CDFSL.
	
	To handle this problem, we notice that \textbf{we can easily find different minima in the representation space}: 
	given an un-normalized input representation, many normalization methods have been proved to be effective in producing discriminative normalized outputs, which can be viewed as different minima in the representation space.
	As the distance between representations is much larger than the sunken region in the loss landscape around each minimum, we can flatten a long-range loss landscape between these minima.
	
	Based on this intuition, we randomly interpolate differently normalized representations at each layer of deep networks, and use interpolated representations for classification. This would push interpolated representations to be effective, lowering the high-loss region between minima and flattening the loss landscape.
	Specifically, we provide two instantiations for both Convolutional Neural Networks and Vision Transformers, and implement the above operations as a normalization layer (FLoR layer) to replace the original normalization layer.
	Our contribution can be listed as 
	
	$\bullet$ To the best of our knowledge, we are the first to extend the analysis of loss landscapes from the parameter space to the representation space for the CDFSL task, which interprets the difficulty in transferring and finetuning.
	
	$\bullet$ Based on the analysis, we propose a simple but effective method to flatten the loss landscape in a long range, which simultaneously enhances the transferring and finetuning of the model.
	
	$\bullet$ We evaluate our model on 8 datasets to show the effectiveness and rationale, indicating we can outperform state-of-the-art works in the average accuracy, and outperform current best works on individual datasets by up to 9\%.

	\section{Analyzing Generalization from the Aspect of Representation-Space Loss Landscapes}
	
	\vspace{-0.1cm}
	\subsection{Preliminaries}
	\vspace{-0.1cm}
	
	Cross-domain few-shot learning (CDFSL) aims to recognize target-domain novel classes ($C^{novel}$) by only a few training samples, with knowledge transferred from source-domain base classes ($C^{base}$) with sufficient training data~\cite{guobroader}. Note that $C^{novel} \cap C^{base} = \varnothing$, and a domain gap exists between these two sets of classes~\cite{guobroader}. 
	The model is firstly trained on the base-class dataset $D^{base} = \{x_i, y_i\}_i^N$ where $y_i \in C^{base}$ (base-class stage).
	Typically, the base-class training is to minimize the cross-entropy loss
	\vspace{-0.1cm}
	{\small
		\begin{equation}
			L = L_{cls}(F(x_i), y_i),
			\label{eq:base_cls}
	\end{equation}}where $F(x_i) = h(f(x_i))$ outputs the classification probability of $x_i$, which is composed of a feature extractor $f(\cdot)$ and a classifier $h(\cdot)$.
	Then, $f(\cdot)$ is transferred to novel-class dataset $D^{novel} = \{x^u_i, y^u_i\}_i^{N^u}$ where $y^u_i \in C^{novel}$ (novel-class stage). Typically, only 1 or 5 training samples are available for each novel class, which makes the training on novel classes difficult.
	For a fair comparison, during the novel-class stage, current works~\cite{guobroader} always sample $k$-way $n$-shot episodes for novel-class training and evaluation. Each episode contains a support set $S = \{x^s_{ij}, y^s_{ij}\}_{i,j}^{k, n}$ with $k$ novel classes and $n$ training samples in each class for training, and a query set $Q = \{x^q_i\}_i^{N^q}$ for evaluation.
	Denote the model trained on novel classes as $F^u(\cdot) = h^u(f^u(\cdot))$, the prediction for $x^q_i$ is 
	\vspace{-0.1cm}
	{\small
		\begin{equation}
			\hat{y}^u_i = \arg max F^u(x^u_i),
			\label{eq:novel_predict}
	\end{equation}}
			Finally, the evaluation can be conducted based on the comparison between $\hat{y}^u_i$ and the real label.
			As can be seen, two major tasks are in CFDSL: (1) obtaining a generalizable feature extractor on source-domain base classes, and (2) effectively finetuning the model given only a few samples on target-domain novel classes.
			To handle these two tasks, we first analyze model's robustness to domain shifts, by expanding the concept of loss-landscape flatness from the parameter space to the representation space.
			
			\vspace{-0.1cm}
			\subsection{Representation-Space Loss Landscape}
			\vspace{-0.1cm}
			
			Given a deep network, its Parameter-Space Loss Landscape (PSLL)~\cite{foret2021sharpnessaware, keskar2017largebatch, zhang2023gradient} refers to drawing a curve to map its parameters to classification losses.
			When the data changes, the model parameters trained on this data will also change, leading to a shift in the loss landscape.
			However, as the data shift is reflected in the parameters trained on the data, it is indirect to see how the domain shift influences the model decision, since the model training is not deterministic.
			To handle this problem, we try to directly analyze how the shift in pixels or representations would influence the model decision (Representation-Space Loss Landscape, RSLL). 
			
			Specifically, denote a data point as $x$, it can be represented in the pixel space (e.g., RGB) or the feature space through a feature extractor, which we uniformly term as the representation space. Denote its representation as $r(x) \in R^{c \times h \times w}$. Since each point in the representation space can be mapped to a loss value, we can draw a landscape corresponding to each representation point and its loss value (black curve in Fig.~\ref{fig:motivation}a).
			During the base-class stage, the model is encouraged to learn a good representation that corresponds to a low classification loss, which can be understood as a minimum in the landscape (black dot in Fig.~\ref{fig:motivation}a).
			After training, each representation extracted by the model from each training sample can be viewed as a minimum.
			
			Then, we fix and transfer this model to another data point $x^u$ with domain shifts against $x$.
			To simplify the analysis, suppose $x^u$ is directly generated by only shifting the domain information of $x$ (Fig.~\ref{fig:motivation}a, the case of different images will be included later).
			Ideally, the model should also map $x^u$ to the same representation point as the original data $x$, since ideally, the model is robust to domain shift.
			However, $x^u$ is likely to be mapped to another point (black cross in Fig.~\ref{fig:motivation}a, $r(x^u)$). Since the model is fixed, the loss landscape is still the original one. Therefore, the mapped point $r(x^u)$ may correspond to a high loss, which means the representation $r(x^u)$ is not effective enough. 
			If the original representation $r(x)$ is located in a sharp minimum, the shifted representation $r(x^u)$ will tend to be located in a high-loss region, demonstrating a badly transferred model.
			Therefore, we can flatten the landscape around the minimum in RSLL. \textbf{It would make the shifted representation located in the same low-loss region as the non-shifted representation, facilitating the model in tolerating larger representation shifts and making the transferring easier (Fig.~\ref{fig:motivation}c).}

			If we finetune the model on the target domain, the model will be guided to minimize the loss of the representation, which will shift both the loss landscape (move the black curve to the blue curve in Fig.~\ref{fig:motivation}a) and the representation point\footnote{If the representation is in the pixel space, this point will not move.} (move the black cross to the blue star in Fig.~\ref{fig:motivation}a) to obtain a new minimum (blue star in Fig.~\ref{fig:motivation}a). 
			However, if the original representation $r(x)$ is located in a sharp minimum, there may be high-loss regions between $r(x)$ and $r(x^u)$. This would make it hard to move $r(x^u)$  to a low-loss region, especially for the few-shot scenarios, making the few-shot finetuning difficult.
			Therefore, we can also flatten the landscape around the minimum in RSLL. \textbf{It would reduce the high-loss region between the source and target representations, making the finetuning easier (Fig.~\ref{fig:motivation}c).}
			
			For the case of different images, there are two gaps between $x$ and $x^u$: (1) the domain shift and (2) the semantic shift. Ideally, we can construct an intermediate sample $x^I$ to represent $x^u$ in the same domain of $x$ but preserve the semantics of $x^u$. If the source domain training set of $x$ is large enough, the model (e.g., Large Vision Models) trained on it will also ensure a good representation for $x^I$. Therefore, this case will come back to the case we analyzed above.

			\begin{figure}[t]
				\centering
				\includegraphics[width=1.0\columnwidth]{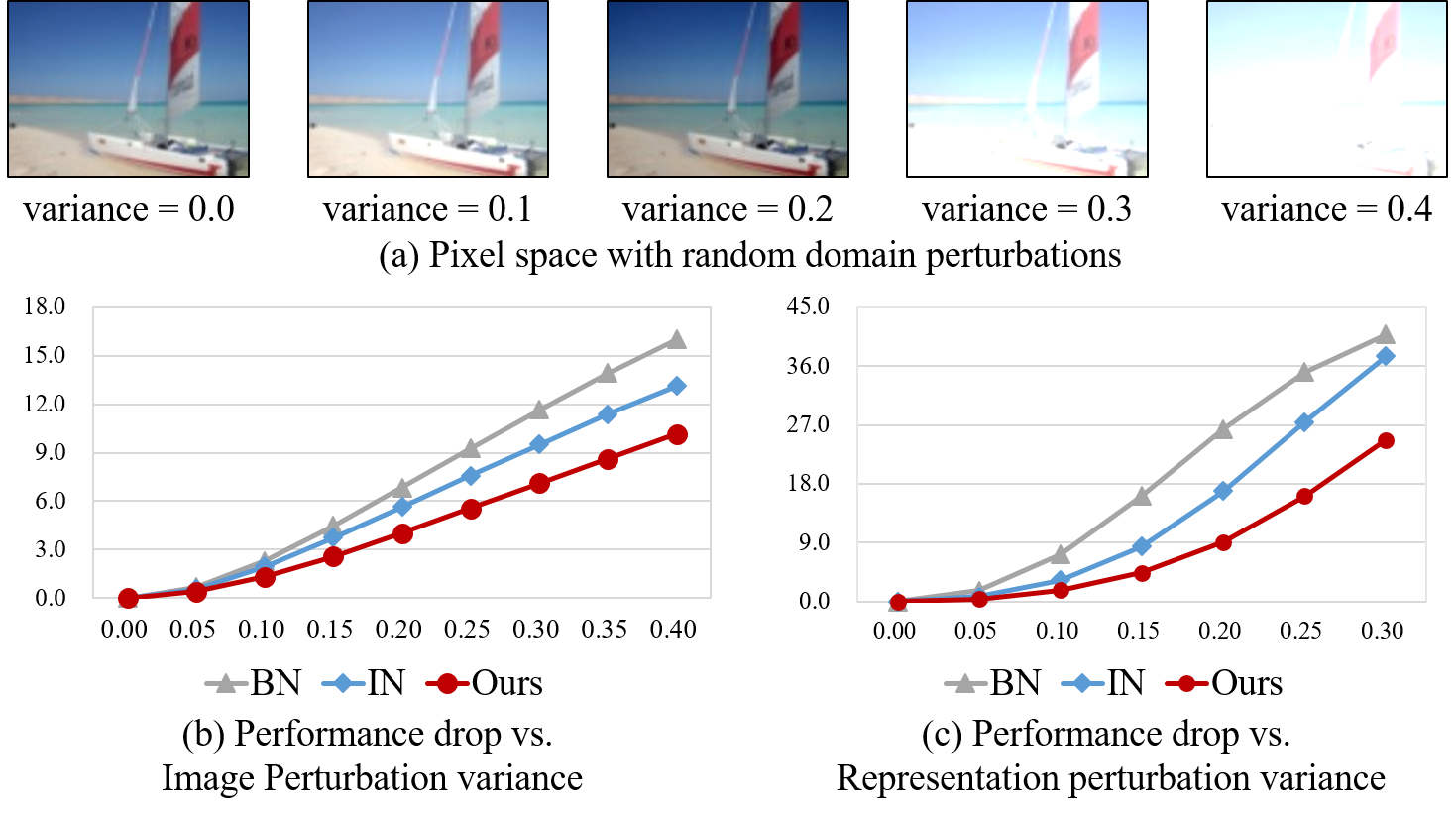}\vspace{-0.2cm}
				\caption{
					To validate the analysis based on representation-space loss landscapes (RSLL), we apply low-frequency noises to the representation space (i.e., pixels and features) to be the domain shifts on training data for a sanity check. 
					The perturbation variance measures the distance between the perturbed representation and the original representation (a minimum in RSLL). We use the performance drop against perturbation variance to measure the sharpness of the landscapes around the minimum, where a larger drop indicates a sharper minimum. 
					We can see the model based on Instance Normalization (IN) is located in a flatter minimum than the model based on Batch Normalization (BN), which brings the high performance of the IN-based model in cross-domain tasks. This result is consistent with current works~\cite{fu2023styleadv} and validates the rationale of the RSLL analysis.
					(a) Samples of the pixel perturbation. (b) Perturbation on pixels. (c) Perturbation on features.
				}\vspace{-0.4cm}
				\label{fig:plot_input_noise_train}
			\end{figure}
			
			In summary, the benefits of RSLL analysis are in
			
			\noindent $\bullet$ It handles domain shifts more directly than PSLL.
			
			\noindent $\bullet$ It interprets the difficulty of transferring and finetuning.
			
			\noindent $\bullet$ We can easily find many minima in RSLL (shown later).
			
			\vspace{-0.1cm}
			\subsection{Verification and Interpretation}
			\vspace{-0.1cm}
			
			To verify the rationale of the above analysis, we first experiment with the influence of domain shifts in the representation space for a sanity check.
			Since the Batch Normalization~\cite{ioffe2015batch} is frequently applied in the widely adopted backbone network for CDFSL, and Instance Normalization~\cite{ulyanov2017instance} has been validated to be beneficial for cross-domain representations~\cite{fu2023styleadv}, we compare these the network with these two networks to validate the rationale of the above analysis. 
			Due to the difficulty in plotting the loss landscape in the high dimensional representation space, we perturb the representation of each data to see the performance drop against the perturbation variance. This variance measures the distance between the perturbed representation and the original representation (a minimum in RSLL). A larger drop indicates a sharper minimum in the landscape. 
			
			We first apply pixel perturbation (i.e., representing each data point by RGB pixels) to the training data. 
			Current works~\cite{fu2022wavesan} show that the domain information is always correlated with the low-frequency information in the Fourier-transformed images. Therefore, we randomly sample a low-frequency perturbation from the Gaussian Distribution, where the Variance controls the distance between the original image and the perturbed image.
			The perturbation is at the shape of $h^p \times w^p$ where $h^p$ and $w^p$ are much smaller than the image size to ensure the frequency is low. Then, the perturbation will be resized to the size of each image and finally applied to them.
			Examples of such perturbation are shown in Fig.~\ref{fig:plot_input_noise_train}a, and the magnitude of the performance drop is shown in Fig.~\ref{fig:plot_input_noise_train}b.
			As can be seen, the performance of all models drops under the perturbations, but the IN model drops less than the BN model, indicating its robustness to the domain shift. 
			This result is consistent with current works~\cite{fu2023styleadv}, and verifies the IN model makes training images located in a flatter minimum in RSLL
			
			\begin{table}[t]
				\begin{center}
					\includegraphics[width=1.0\columnwidth]{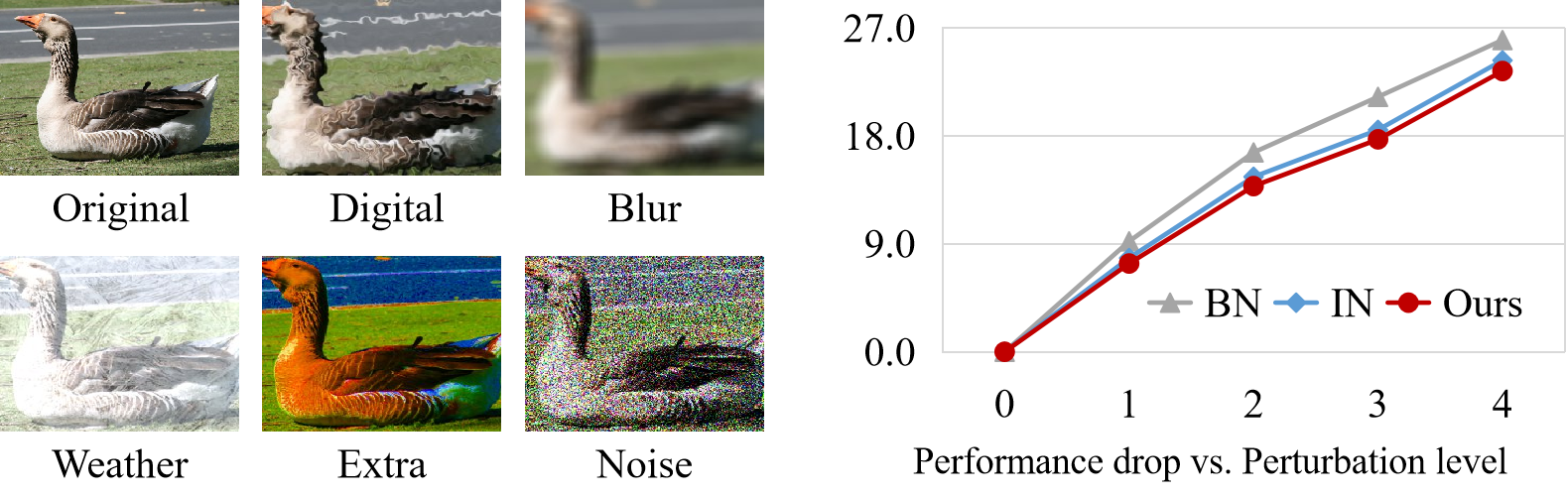}\\
					\resizebox{0.48\textwidth}{!}{
						\begin{tabular}{ccccccc}
							\toprule
							Method & Not Shifted & Digital & Blur & Extra & Noise & Weather \\
							\midrule
							BN & 62.61 {\scriptsize$\pm$0.18} & 44.98 {\scriptsize$\pm$0.16} & 41.98 {\scriptsize$\pm$0.15} & 37.22 {\scriptsize$\pm$0.13} & 40.08 {\scriptsize$\pm$0.14} & 33.53 {\scriptsize$\pm$0.12} \\
							IN & 68.70 {\scriptsize$\pm$0.18} & 54.92 {\scriptsize$\pm$0.18} & 49.92 {\scriptsize$\pm$0.18} & 44.30 {\scriptsize$\pm$0.16} & 47.56 {\scriptsize$\pm$0.16} & 40.61 {\scriptsize$\pm$0.14} \\
							Ours & \textbf{68.89} {\scriptsize$\pm$0.18} & \textbf{55.44} {\scriptsize$\pm$0.18} & \textbf{50.63} {\scriptsize$\pm$0.17} & \textbf{45.74} {\scriptsize$\pm$0.16} & \textbf{48.80} {\scriptsize$\pm$0.16} & \textbf{42.29} {\scriptsize$\pm$0.14} \\
							\bottomrule
					\end{tabular}}\vspace{-0.2cm}
					\caption{
						To verify our analysis also holds for the real-world data,
						we evaluate our model on the test data of domain shifted \textit{mini}ImageNet by 5-way 1-shot accuracy. Datasets are sampled from the ImageNet-c dataset with 5 kinds of shifts.
						This dataset provides 5 levels of perturbations, so we use the perturbation level to measure how the representation shifts from the original representation. The sharpness is also measured by the performance drop. We can see IN-based model still shows a flatter minimum in RSLL, which leads to its high performance in these datasets.
					}
					\label{tab:mini-c}
				\end{center}\vspace{-0.45cm}
			\end{table}
			
			Then, we apply perturbations to feature maps (i.e., represent each sample in the feature space). The perturbation is also the low-frequency Gaussian noise, and the performance drop is in Fig.~\ref{fig:plot_input_noise_train}c, where we can see similar results as in Fig.~\ref{fig:plot_input_noise_train}b.
			This result verifies the RSLL analysis also fits the feature map representation for each input sample.
			
			To verify this analysis applies to real-world domain shifts, we then construct 5 domain-shifted \textit{mini}ImageNet by selecting the same classes from ImageNet-c~\cite{hendrycks2019benchmarking}. The domain shifts contain blur, digital, extra, noise, and weather perturbations. Examples and results can be found in Tab.~\ref{tab:mini-c}.
			This dataset provides 5 levels of perturbations for each kind of perturbation, so we utilize this level to represent the distance between the perturbed representation and the original one (a minimum in the RSLL). Similar to Fig.~\ref{fig:plot_input_noise_train}, the sharpness is measured by the performance drop.
			We can see the results are consistent with the toy experiments in Fig.~\ref{fig:plot_input_noise_train}, which further validates the rationale of the RSLL analysis.
			
			\begin{figure}[t]
				\begin{center}
					\includegraphics[width=1.0\columnwidth]{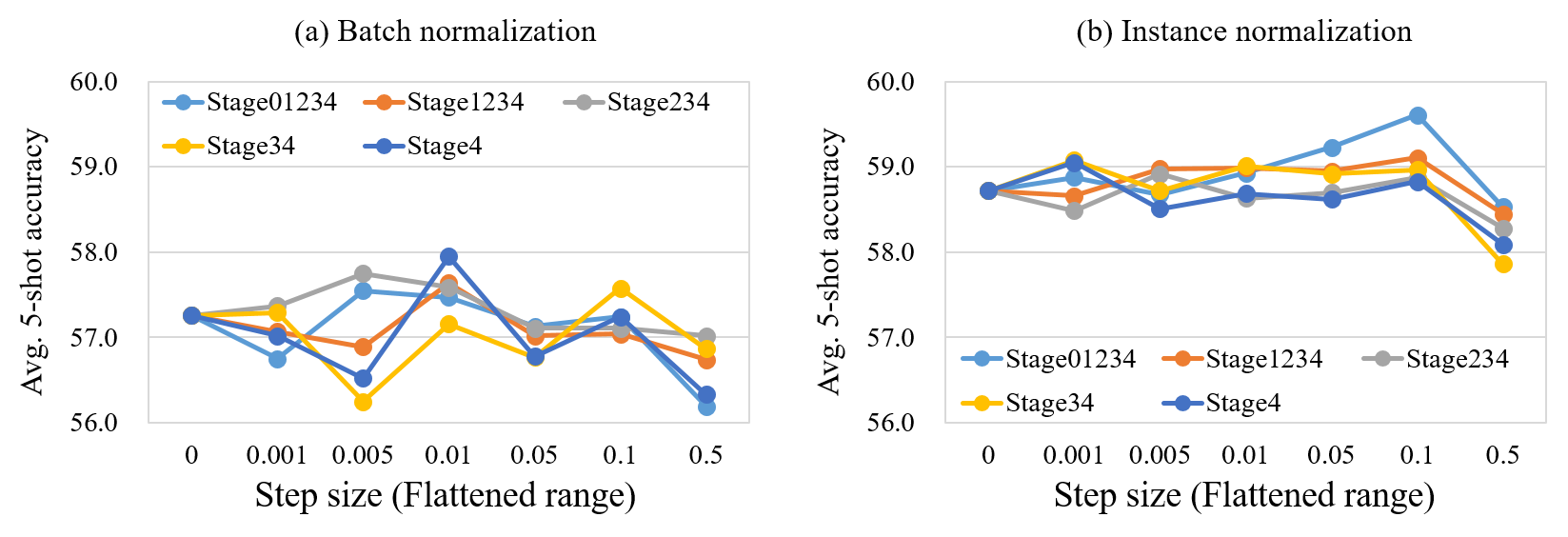}\vspace{-0.3cm}
					\caption{Directly apply SAM on each representation of ResNet10 on CDFSL datasets. Only marginal improvements on the average CDFSL 5-way 5-shot accuracy can be observed, and the perturbation step size is small, indicating the complex loss landscape could only support SAM to learn a short-range flatness.}
					\label{fig:sam_on_rep}
				\end{center}\vspace{-0.45cm}
			\end{figure}
			
			\begin{table}[t]
				\begin{center}
					\caption{Our perturbation step size (flattened range) is much larger than SAM in the representation output by each layer, no matter when it is at the beginning (Init.) or the end of the training. }\vspace{-0.3cm}
					\label{tab:IN_BN_dis_h}
					\hspace*{-0.3cm}
					\resizebox{0.5\textwidth}{!}{
						\begin{tabular}{ccccccccccccc}
							\toprule
							Rep. ID & 1 & 2 & 3 & 4 & 5 & 6 & 7 & 8 & 9 & 10 & 11 & 12 \\
							\midrule
							Init. & 5.06 & 3.45 & 2.12 & 3.58 & 1.99 & 3.57 & 3.25 & 2.33 & 3.34 & 4.69 & 3.80 & 4.84 \\
							End. & 2.91 & 2.44 & 1.51 & 2.60 & 2.09 & 1.94 & 3.54 & 3.05 & 1.80 & 5.69 & 12.46 & 9.18\\
							\bottomrule
					\end{tabular}}
				\end{center}\vspace{-0.45cm}
			\end{table}
			
			In summary, by comparing IN and BN models on domain-shifted datasets, experiments in Fig.~\ref{fig:plot_input_noise_train} and Tab.~\ref{tab:mini-c} validate (1) the RSLL analysis is rationale; (2) the domain perturbation on pixels and features, which we uniformly name as representation, are consistent. Moreover, it also interprets why IN models are better, as IN models' representations are in flatter minima compared with BN models.
			
			\vspace{-0.1cm}
			\subsection{Short-Range Flatness Limits Generalization}
			\vspace{-0.1cm}
			
			To improve the model's robustness to domain shifts, current works~\cite{foret2021sharpnessaware} provided an effective method named Sharpness-Aware Minimization (SAM). It applies adversarial perturbation to the model's parameters to flatten the PSLL. It inspires us to conduct SAM on the RSLL. Take the last layer's representation as an example, the training loss is
			{\small
				\begin{equation}
					L = L_{cls}(h(f(x) + \eta \frac{\nabla f(x)}{||\nabla f(x)||} ), y),
					\label{eq:sam_on_rep}
			\end{equation}}where $\eta$ is the step size to control the flattened range, $x$ is the input sample, $y$ is the label, and $\nabla f(x) = \frac{\partial L}{\partial f(x)}$. Results are reported in Fig.~\ref{fig:sam_on_rep}, including the step size $\eta$ and perturbed layers. However, we can only see marginal improvements against the baseline performance ($\eta=0$). 
			
			Moreover, we can observe the step sizes are small, but the IN-based model could tolerate a larger step size with a higher performance. Combining this result with the results in Fig.~\ref{fig:plot_input_noise_train} and Tab.~\ref{tab:mini-c}, we hold that this is because the IN-based model extracts features in a flatter RSLL minimum. 
			As the perturbation direction is obtained by $\nabla f(x_i)$ which only estimates the local shape of RSLL, when the minimum is sharp, it fails to estimate the complex loss landscape shape outside the local minimum of $f(x_i)$. Therefore, the perturbation direction $\nabla f(x_i)$ will be less effective when the step size is large.
			In contrast, since IN-based representations are in flatter minima than BN-based ones, it ensures $\nabla f(x_i)$ is still effective given a larger perturbation step size. In this scenario, a larger step size would help the model to flat a longer range of RSLL and improve generalization.
			
			Therefore, to learn a model robust to domain shifts, we aim to flatten a long-range RSLL around each representation. Our flattened range is reported in Tab.~\ref{tab:IN_BN_dis_h} on the representation output by each layer, where the range is much larger than ordinary SAM methods in Fig.~\ref{fig:sam_on_rep}.

			\vspace{-0.1cm}
			\section{Flatten Long-Range Loss Landscapes}
			\vspace{-0.1cm}
			
			\begin{figure}[t]
				\centering
				\includegraphics[width=1.0\columnwidth]{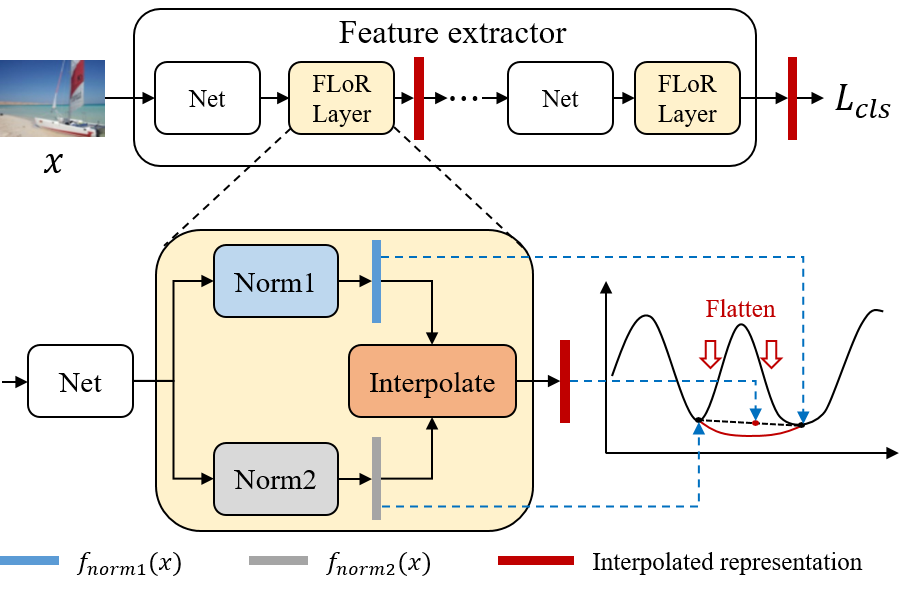}\vspace{-0.4cm}
				\caption{We implement our method as a normalization layer (FLoR layer) to replace the ordinary normalization layer in the backbone network (e.g., CNNs or ViTs). This layer interpolates two differently normalized representations, which flattens the intermediate high-loss region between two minima in RSLL.}\vspace{-0.4cm}
				\label{fig:framework}
			\end{figure}
			
			To enlarge the range of flattened area, basically relying on the local information (such as gradient~\cite{foret2021sharpnessaware} or curvature~\cite{zhang2023gradient}) of the optimum may not be enough, because this information is only effective inside the sunken area in the landscape. 
			
			To handle this problem, we notice a distinguished difference between the PSLL and RSLL: \textbf{there are many easy-to-find minima in RSLL}. For example, previous works proposed many different normalization methods, such as Batch Normalization or Instance Normalization. These methods can provide effective representations suitable for classification, although they may show differences in the fitness for different tasks.
			In other words, given the same training sample, different normalization methods could provide different minima in the same representation-space loss landscape. This inspires us to flatten the loss in the area between these minima (Fig.~\ref{fig:motivation}b).
			
			Specifically, we utilize two normalization methods to normalize each pre-normalized representation, which produces two effective representations and corresponds to two minima in RSLL. Then, we propose to flatten the loss landscapes between these two minima, by means of classifying the input sample through the interpolated representations between two minima. Take the final layer's representation $f(x)$ as an example, the classification can be represented as 
			\vspace{-0.1cm}
			{\small
				\begin{equation}
					L = L_{cls}(h((1 - \delta) f_{norm1}(x) + \delta f_{norm2}(x)), y),
					\label{eq:final_loss}
			\end{equation}}where $f_{norm1}(x)$ and $f_{norm2}(x)$ refer to two normalized representations, and $\delta \in [0, 1]$ is a ratio randomly sampled from the Beta distribution. In implementation, we conduct the interpolation in \textbf{every} layer's representation.
			Since the classification is based on the interpolated representation, this representation will be pushed to be effective, so it will be mapped to a low loss in the RSLL. Thus, the high-loss region between $f_{norm1}(x)$ and $f_{norm2}(x)$ will be flattened.
			
			As this method is agnostic to the shape of the complex loss landscape between minima, we do not need to consider any local information around minima. We measure the distance between the BN and IN representations in Tab.~\ref{tab:IN_BN_dis_h}, where the distance of each representation is much larger than step sizes in Fig.~\ref{fig:sam_on_rep}, indicating a larger flattened region.
			
			Below, we provide two instantiations of the above idea for both Convolutional Neural Networks (CNN)~\cite{he2016deep} and Vision Transformers (ViT)~\cite{dosovitskiy2021image} respectively.
			
			\vspace{-0.1cm}
			\subsection{Flattening for Convolutional Neural Networks} 
			\vspace{-0.1cm}
			
			In CNNs, Batch Normalization is one of the most widely used normalization methods, which is typically applied after the convolution layer.
			BN refers to normalizing representations with batch statistics, which can be represented as
			{\small
				\begin{align}
					& f_{BN}(X) = \gamma \frac{X - \mu}{\sqrt{\sigma^2 + \epsilon}} + \beta, \\
					& \mu = \frac{1}{bhw} \sum_{i,j,k} X_{i,:,j,k}, \quad \sigma^2 = \frac{1}{bhw} \sum_{i,j,k} (X_{i,:,j,k} - \mu)^2, \nonumber
					\label{eq:BN}
			\end{align}}where $X \in R^{b \times c \times h \times w}$ is the batch representation. 
			
			Instance normalization (IN) is another widely used normalization method for CNNs, which refers to normalizing representations with only the given image statistics as
			{\small
				\begin{align}
					&f_{IN}(X) = \gamma \frac{X - \mu}{\sqrt{\sigma^2 + \epsilon}} + \beta, \\
					&\mu = \frac{1}{hw} \sum_{j,k} X_{:,:,j,k},
					\sigma^2 = \frac{1}{hw} \sum_{j,k} (X_{:,:,j,k} - \mu)^2. \nonumber
					\label{eq:IN}
			\end{align}}
			
			For CNN, we set $f_{norm1}$ to $f_{BN}$, and set $f_{norm2}$ to $f_{IN}$.
			
			\vspace{-0.1cm}
			\subsection{Flattening for Vision Transformers} 
			\vspace{-0.1cm}
			
			In ViTs, Layer Normalization is widely applied as the normalization method,
			which is represented as
			{\small
				\begin{align}
					& f_{LN}(X) = \gamma \frac{X - \mu}{\sqrt{\sigma^2 + \epsilon}} + \beta, \\
					& \mu = \frac{1}{c} \sum_{k} X_{:,:,k}, \quad \sigma^2 = \frac{1}{c} \sum_{:,:,k} (X_{:,:,k} - \mu)^2, \nonumber
					\label{eq:LN}
			\end{align}}where $X \in R^{b \times t \times c}$ is the batch representation.
			
			Then, we follow \cite{chen2023separate} to only process the CLS token. We use BN as the second normalization method. In other words, we set $f_{norm1}$ to $f_{LN}$, and set $f_{norm2}$ to $f_{BN}$ for the CLS token in each layer.
			
			\vspace{-0.1cm}
			\subsection{Implementation}
			\vspace{-0.1cm}
			
				
				In implementation, as shown in Fig.~\ref{fig:framework}, we implement the above design as a normalization layer (FLoR layer) to replace the ordinary normalization layer in the backbone network. This layer can be represented as 
				{\small
					\begin{equation}
						f_{FLL}(x) = (1 - \delta) f_{norm1}(x) + \delta f_{norm2}(x),
						\label{eq:mix}
				\end{equation}}which is the same as Eq.~\ref{eq:final_loss}.
				Since $\delta$ is a random number sampled from the Beta distribution $Beta(a, b)$, our model imports only two hyper-parameters ($a$, $b$) and learnable parameters only in the added normalization layer, which is simple and lightweight.
				
				During the base-class training, $\delta$ is randomly sampled from the $Beta(a, b)$. During the novel-class finetuning (if needed), $\delta$ is set to a learnable parameter that is finetuned together with the whole network, where its value is initialized as the expectation value of $Beta(a, b)$. Finally, we evaluate the model by either the prototype-based classifier or the finetune-based classifier, as illustrated in Eq.~\ref{eq:novel_predict}.
				
				\vspace{-0.1cm}
				\section{Experiments}
				\vspace{-0.1cm}
				
				
				\vspace{-0.1cm}
				\subsection{Dataset and evaluation setup}
				\vspace{-0.1cm}
				
				Following current works~\cite{DBLP:journals/corr/abs-1904-04232,guobroader}, our model is firstly trained on the \textit{mini}ImageNet dataset~\cite{Vinyals2016Matching} (base classes, source domain) and then transferred to 8 cross-domain dataset (novel classes, target domain, including CUB~\cite{wah2011caltech}, Cars~\cite{6755945}, Plantae~\cite{vanhorn2018inaturalist}, Places~\cite{7968387}, CropDiseases~\cite{34699834fa624a3bbc2fae48eb151339}, EuroSAT~\cite{helber2019eurosat}, ISIC2018~\cite{codella2019skin} and ChestX~\cite{Wang_2017}) for few-shot training and evaluation, using the $k$-way $n$-shot classification. Dataset information is listed in Tab.~\ref{tab:dataset}.
				
				\begin{table}[t]
					\begin{center}
						\caption{Dataset information. Please see the appendix for details. }\vspace{-0.3cm}
						\label{tab:dataset}
						\resizebox{0.4\textwidth}{!}{
							\begin{tabular}{cccc}
								\toprule
								Dataset & Domain & Classes & Images \\
								\midrule
								\textit{mini}ImageNet & General recognition & 64 & 38,400 \\
								CUB & Fine-grained bird recognition & 50 & 2,953 \\
								Cars & Fine-grained car recognition & 49 & 2,027 \\
								Plantae & Plantae recognition & 50 & 17,253 \\
								Places & Scene recognition & 19 & 3,800 \\
								CropDiseases & Agricultural disease recognition & 38 & 43,456 \\
								EuroSAT & Satellite imagery recognition & 10 & 27,000 \\
								ISIC2018 & Skin lesion recognition & 7 & 10,015 \\
								ChestX & X-ray chest recognition & 7 & 25,847 \\
								\bottomrule
						\end{tabular}}
					\end{center}\vspace{-0.6cm}
				\end{table}
				
				\vspace{-0.1cm}
				\subsection{Implementation details}
				\vspace{-0.1cm}
				
				During the base-class training, our model is trained with the AdamW optimizer~\cite{loshchilov2019decoupled} for 400 epochs with a learning rate of 0.001. For a fair comparison, we follow current works~\cite{fu2023styleadv} to utilize ResNet10~\cite{he2016deep} and Vit-S~\cite{dosovitskiy2021image} (with DINO~\cite{zhang2022dino} pretraining on ImageNet1K~\cite{deng2009imagenet}) as the backbone network. We set the Beta distribution as $a=b=0.01$ for all experiments.
				During the novel-class finetuning, we use the SGD optimizer with a momentum of 0.9 to finetune all parameters. 
				Please see the appendix for details.
				
				\vspace{-0.1cm}
				\subsection{Comparison with state-of-the-art works}
				\vspace{-0.1cm}
				
				Comparisons utilizing ResNet10 are listed in Tab.~\ref{tab:sota_1shot} and \ref{tab:sota_5shot}.
				For fairness, we group work by whether finetuning (FT) or the transductive (TR) setting is used. For all groups, we follow MN to use the cosine similarity, and subtract the mean of the support (query) set for inductive (transductive) calibration. For the transductive setting, we pseudo-label the query set by the prototype-based classifier or the finetune-based classifier, and use the pseudo-labeled query set to further update the classifier until convergence.
				As shown in Tab.~\ref{tab:sota_1shot} and \ref{tab:sota_5shot}, we achieve the top average performance in all settings, and achieve the highest performance in almost all datasets. Notably, on ISIC2018, we can significantly outperform state-of-the-art works by as much as 9\% in the 5-shot setting. 
				Comparisons with works using ViT-S are in Tab.~\ref{tab:sota_vit}, and we also achieve state-of-the-art performance.

				\begin{table*}[t]
					\begin{center}
						\caption{Comparison with state-of-the-art works based on ResNet10 by 5-way 1-shot accuracy. }\vspace{-0.3cm}
						\label{tab:sota_1shot}
						\resizebox{0.75\textwidth}{!}{
							\begin{tabular}{ccccccccccccc}
								\toprule
								Method & FT & TR & Mark & CUB & Cars & Places & Plantae & ChestX & ISIC2018 & EuroSAT & CropDiseases & Average \\
								\midrule 
								GNN+FT~\cite{tseng2020cross} & $\times$ & $\times$ & ICLR-20 & 45.50 & 32.25 & 53.44 & 32.56 & 22.00 & 30.22 & 55.53 & 60.74 & 51.53 \\
								GNN+ATA~\cite{wang2021crossdomain} & $\times$ & $\times$ & IJCAI-21 & 45.00 & 33.61 & 53.57 & 34.42 & 22.10 & 33.21 & 61.35 & 67.47 & 43.84 \\
								MN+AFA~\cite{hu2022adversarial} & $\times$ & $\times$ & ECCV-22 & 41.02 & 33.52 & \textbf{54.66} & 37.60 & 22.11 & 32.32 & 61.28 & 60.71 & 42.90 \\
								GNN+AFA~\cite{hu2022adversarial} & $\times$ & $\times$ & ECCV-22 & 46.86 & 34.25 & 54.04 & 36.76 & 22.92 & 33.21 & 63.12 & 67.61 & 44.85 \\
								LDP-net~\cite{Zhou_2023_CVPR} & $\times$ & $\times$ & CVPR-23 & 49.82 & 35.51 & 53.82 & 39.84 & 23.01 & 33.97 & \textbf{65.11} & 69.64 & 46.34 \\
								\textbf{FLoR} &$\times$ & $\times$ & \textbf{Ours} & \textbf{49.99} & \textbf{37.41} & 53.18 & \textbf{40.10} & \textbf{23.11} & \textbf{38.11} & 62.90 & \textbf{73.64} & \textbf{47.31} \\
								\midrule
								Fine-tuning~\cite{guobroader}  & $\checkmark$ & $\times$ & ECCV-20 & 43.53 & 35.12 & 50.57 & 38.77 & 22.13 & 34.60 & 66.17 & 73.43 & 45.54 \\
								\textbf{FLoR}  & $\checkmark$ & $\times$ & \textbf{Ours} & \textbf{50.01} & \textbf{38.13} & \textbf{53.61} & \textbf{40.20} & \textbf{23.12} & \textbf{38.81} & \textbf{69.13} & \textbf{84.04} & \textbf{49.63} \\
								\midrule
								TPN+ATA~\cite{wang2021crossdomain} & $\times$ & $\checkmark$ & IJCAI-21 & 50.26 & 34.18 & 57.03 & 39.83 & 21.67 & 34.70 & 65.94 & 77.82 & 47.68 \\
								TPN+AFA~\cite{hu2022adversarial} & $\times$ & $\checkmark$ & ECCV-22 & 50.85 & 38.43 & 60.29 & 40.27 & 21.69 & 34.25 & 66.17 & 72.44 & 48.05 \\
								\textbf{FLoR} & $\times$ & $\checkmark$ & \textbf{Ours} & \textbf{55.35} & \textbf{38.86} & \textbf{60.94} & \textbf{41.61} & \textbf{22.92} & \textbf{39.78} & \textbf{70.96} & \textbf{85.95} & \textbf{52.04} \\
								\midrule
								TPN+ATA~\cite{wang2021crossdomain} & $\checkmark$ & $\checkmark$ & IJCAI-21 & 51.89 & 38.07 & 57.26 & 40.75 & 22.45 & 35.55 & 70.84 & 82.47 & 49.91 \\
								RDC~\cite{li2022ranking} & $\checkmark$ & $\checkmark$ & CVPR-22 & 50.09 & 39.94 & 61.17 & 41.30 & 22.32 & 36.28 & 70.51 & 85.79 & 50.81 \\
								LDP-net~\cite{Zhou_2023_CVPR} & $\checkmark$ & $\checkmark$ & CVPR-23 & 55.94 & 37.44 & \textbf{62.21} & 41.04 & 22.21 & 33.44 & \textbf{73.25} & 81.24 & 50.85 \\
								\textbf{FLoR}  & $\checkmark$ & $\checkmark$ & \textbf{Ours} & \textbf{55.94} & \textbf{40.01} & 61.27 & \textbf{41.70} & \textbf{23.12} & \textbf{41.67} & 71.38 & \textbf{86.30} & \textbf{52.67} \\
								\bottomrule
						\end{tabular}}
					\end{center}\vspace{-0.6cm}
				\end{table*}
				
				\vspace{-0.1cm}
				\subsection{Ablation study}
				\vspace{-0.1cm}
				
				\subsubsection{Verification of local loss landscapes}
				\vspace{-0.1cm}

				\begin{table*}[t]
					\begin{center}
						\caption{Comparison with state-of-the-art works based on ResNet10 by 5-way 5-shot accuracy. }\vspace{-0.3cm}
						\label{tab:sota_5shot}
						\resizebox{0.75\textwidth}{!}{
							\begin{tabular}{ccccccccccccc}
								\toprule
								Method & FT & TR & Mark & CUB & Cars & Places & Plantae & ChestX & ISIC2018 & EuroSAT & CropDiseases  & Average \\
								\midrule 
								GNN+FT~\cite{tseng2020cross} & $\times$ & $\times$ & ICLR-20 & 64.97 & 46.19 & 70.70 & 49.66 & 24.28 & 40.87 & 78.02 & 87.07 & 57.72 \\
								GNN+ATA~\cite{wang2021crossdomain} & $\times$ & $\times$ & IJCAI-21 & 66.22 & 49.14 & 75.48 & 52.69 & 24.32 & 44.91 & 83.75 & 90.59 & 60.89 \\
								MN+AFA~\cite{hu2022adversarial} & $\times$ & $\times$ & ECCV-22 & 59.46 & 46.13 & 68.87 & 52.43 & 23.18 & 39.88 & 69.63 & 80.07 & 54.96 \\
								GNN+AFA~\cite{hu2022adversarial} & $\times$ & $\times$ & ECCV-22 & 68.25 & 49.28 & \textbf{76.21} & 54.26 & 25.02 & 46.01 & \textbf{85.58} & 88.06 & 61.58 \\
								LDP-net~\cite{Zhou_2023_CVPR} & $\times$ & $\times$ & CVPR-23 & 70.39 & 52.84 & 72.90 & \textbf{58.49} & 26.67 & 48.06 & 82.01 & 89.40 & 62.60 \\
								\textbf{FLoR}& $\times$ & $\times$ & \textbf{Ours} & \textbf{70.39} & \textbf{53.43} & 72.31 & 55.80 & \textbf{26.70} & \textbf{51.44} & 80.87 & \textbf{91.25} & \textbf{62.77} \\
								\midrule
								Fine-tuning~\cite{guobroader}  & $\checkmark$ & $\times$ & ECCV-20 & 63.76 & 51.21 & 70.68 & 56.45 & 25.37 & 49.51 & 81.59 & 89.84 & 61.05 \\
								\textbf{FLoR} & $\checkmark$ & $\times$ & \textbf{Ours} & \textbf{73.39} & \textbf{57.21} & \textbf{72.37} & \textbf{61.11} & \textbf{26.77} & \textbf{56.74} & \textbf{83.06} & \textbf{92.33} & \textbf{65.37} \\
								\midrule
								TPN+ATA~\cite{wang2021crossdomain} & $\times$ & $\checkmark$ & IJCAI-21 & 65.31 & 46.95 & 72.12 & 55.08 & 23.60 & 45.83 & 79.47 & 88.15 & 59.56 \\
								TPN+AFA~\cite{hu2022adversarial} & $\times$ & $\checkmark$ & ECCV-22 & 65.86 & 47.89 & 72.81 & 55.67 & 23.47 & 46.29 & 80.12 & 85.69 & 59.73 \\
								\textbf{FLoR} & $\times$ & $\checkmark$ & \textbf{Ours} & \textbf{70.83} & \textbf{53.55} & \textbf{73.88} & \textbf{56.28} & \textbf{26.27} & \textbf{52.16} & \textbf{82.04} & \textbf{92.32} & \textbf{63.42} \\
								\midrule
								TPN+ATA~\cite{wang2021crossdomain} & $\checkmark$ & $\checkmark$ & IJCAI-21 & 70.14 & 55.23 & 73.87 & 59.02 & 24.74 & 49.83 & \textbf{85.47} & 93.56 & 63.98 \\
								RDC~\cite{li2022ranking} & $\checkmark$ & $\checkmark$ & CVPR-22 & 67.23 & 53.49 & 74.91 & 57.47 & 25.07 & 49.91 & 84.29 & 93.30 & 63.21 \\
								LDP-net~\cite{Zhou_2023_CVPR} & $\checkmark$ & $\checkmark$ & CVPR-23 & 73.34 & 53.06 & \textbf{75.47} & 59.64 & 26.88 & 48.44 & 84.05 & 91.89 & 64.10 \\
								\textbf{FLoR} & $\checkmark$ & $\checkmark$ & \textbf{Ours} & \textbf{74.06} & \textbf{57.98} & 74.25 & \textbf{61.70} & \textbf{26.89} & \textbf{57.54} & 83.76 & \textbf{93.60} & \textbf{66.22} \\
								\bottomrule
						\end{tabular}}
					\end{center}\vspace{-0.7cm}
				\end{table*}
				
				\begin{table*}[t]
					\begin{center}
						\caption{Comparison with state-of-the-art works with VIT-S pretrained by DINO on ImageNet-1K. }\vspace{-0.3cm}
						\label{tab:sota_vit}
						\resizebox{0.75\textwidth}{!}{
							\begin{tabular}{ccccccccccccc}
								\toprule
								Method & Shot & FT & Mark & CUB & Cars & Places & Plantae & ChestX & ISIC2018 & EuroSAT & CropDiseases  & Average \\
								\midrule 
								StyleAdv~\cite{fu2023styleadv} & 1 & $\times$ & CVPR-23 & 84.01 & 40.48 & 72.64 & \textbf{55.52} & \textbf{22.92} & 33.05 & 72.15 & 81.22 & 57.75 \\
								\textbf{FLoR} & 1 & $\times$ & \textbf{Ours} & \textbf{84.60} & \textbf{40.71} & \textbf{73.85} & 51.93 & 22.78 & \textbf{34.20} & \textbf{72.39} & \textbf{81.81} & \textbf{57.78} \\
								\midrule
								PMF~\cite{Shell2023Pushing} & 1 & $\checkmark$ & CVPR-22 & 78.13 & 37.24 & 71.11 & 53.60 & 21.73 & 30.36 & 70.74 & 80.79 & 55.46 \\
								StyleAdv~\cite{fu2023styleadv} & 1 & $\checkmark$ & CVPR-23 & 84.01 & 40.48 & 72.64 & \textbf{55.52} & 22.92 & 33.99 & \textbf{74.93} & \textbf{84.11} & 58.57 \\
								\textbf{FLoR} & 1 & $\checkmark$ & \textbf{Ours} & \textbf{85.40} & \textbf{43.42} & \textbf{74.69} & 52.29 & \textbf{23.26} & \textbf{35.49} & 73.09 & 83.55 & \textbf{58.90} \\
								\midrule
								StyleAdv~\cite{fu2023styleadv} & 5 & $\times$ & CVPR-23 & 95.82 & 61.73 & 88.33 & \textbf{75.55} & \textbf{26.97} & 47.73 & 88.57 & 94.85 & 72.44 \\
								\textbf{FLoR} & 5 & $\times$ & \textbf{Ours} & \textbf{96.1}8 & \textbf{61.75} & \textbf{89.23} & 72.80 & 26.71 & \textbf{49.52} & \textbf{90.41} & \textbf{95.28} & \textbf{72.74} \\
								\midrule
								PMF~\cite{Shell2023Pushing} & 5 & $\checkmark$ & CVPR-22 & - & - & - & - & \textbf{27.27} & 50.12 & 85.98 & 92.96 & - \\
								StyleAdv~\cite{fu2023styleadv} & 5 & $\checkmark$ & CVPR-23 & 95.82 & 66.02 & 88.33 & \textbf{78.01} & 26.97 & 51.23 & 90.12 & 95.99 & 74.06 \\
								\textbf{FLoR} & 5 & $\checkmark$ & \textbf{Ours} & \textbf{96.53} & \textbf{68.44} & \textbf{89.48} & 76.22 & 27.02 & \textbf{53.06} & \textbf{90.75} & \textbf{96.47} & \textbf{74.75} \\
								\bottomrule
						\end{tabular}}
					\end{center}\vspace{-0.7cm}
				\end{table*}
				
\begin{figure}[t]
					\centering
					\includegraphics[width=1.0\columnwidth]{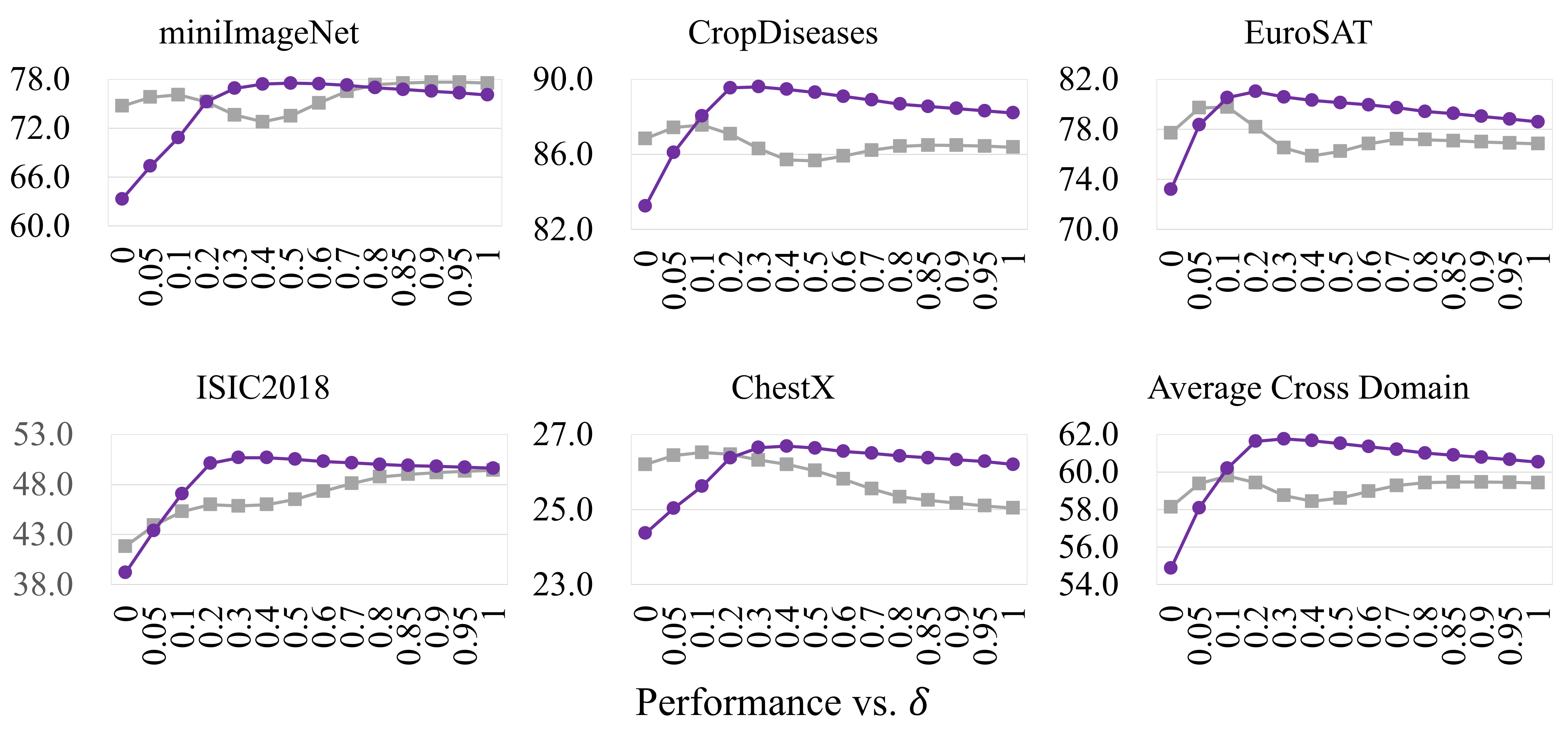}\vspace{-0.3cm}
					\caption{To verify the loss landscape between BN and IN representations, we train a baseline model with separate streams of IN and BN, by means of specifying different mixing ratios $\delta$. Two peaks and one valley can be observed in the baseline curve (gray), indicating the high-loss region between the two representations. In contrast, only one peak is observed in our curve (purple), indicating we can effectively flatten the loss landscapes.}\vspace{-0.5cm}
					\label{fig:plot_inW_before+after}
				\end{figure}

				To verify the loss landscape between two normalized representations, 
				intermediate points are needed to be sampled between these two representations. 
				We first train a network with a shared convolution layer and two separate normalizations in each convolution-normalization block. During training, each branch's final output will be classified separately. 
				During the evaluation, we manually mix two representations in each layer of this model by specifying different $\delta$ in Eq.~\ref{eq:mix}. Since the loss value is not consistent for comparison between different models, we measure the 5-way 5-shot accuracy of the plain prototype-based classification results on different domains, which is plotted in Fig.~\ref{fig:plot_inW_before+after} as the gray curves.
				We can see that almost all datasets show two peaks and one valley in accuracy, verifying the existence of the high-loss region between the two representations.
				
				Similarly, we also manually specify different $\delta$ to form a mixed representation for our model, and measure the corresponding accuracy (purple curve in Fig.~\ref{fig:plot_inW_before+after}). 
				We can see (1) our model achieves higher performance, especially on cross-domain datasets, (2) there is only one peak in the curve and (3) the high-accuracy (low-loss) region is much larger than that of the baseline model, verifying the flatness.
				
				\vspace{-0.1cm}
				\subsubsection{Verification of each design}
				\vspace{-0.1cm}
				
				\begin{table}[t]
					\begin{center}
						\caption{Ablation study of RSLL, Mixup, IN, perturbations and $\delta$ by 5-way 5-shot accuracy with plain prototype classifier. }\vspace{-0.3cm}
						\label{tab:ablation}
						\resizebox{0.48\textwidth}{!}{
							\begin{tabular}{cccccc}
								\toprule
								Method  & CropDisease & EuroSAT & ISIC2018 & ChestX & Ave. \\
								\midrule
								BN (Baseline) & 85.80 {\scriptsize$\pm$0.27} & 78.01 {\scriptsize$\pm$0.22} & 39.10 {\scriptsize$\pm$0.33} & 26.13 {\scriptsize$\pm$0.17} & 57.26 {\scriptsize$\pm$0.13} \\
								PSLL SAM & 86.60 {\scriptsize$\pm$0.44} & 76.83 {\scriptsize$\pm$0.50} & 40.42 {\scriptsize$\pm$0.42} & 26.03 {\scriptsize$\pm$0.34} & 57.47 {\scriptsize$\pm$0.38} \\
								RSLL SAM & 86.19 {\scriptsize$\pm$0.32} & 78.03 {\scriptsize$\pm$0.33} & 41.26 {\scriptsize$\pm$0.31} & 26.26 {\scriptsize$\pm$0.17} & 57.95 {\scriptsize$\pm$0.17} \\
								Mixup & 86.78 {\scriptsize$\pm$0.33} & 78.49 {\scriptsize$\pm$0.31} & 42.89 {\scriptsize$\pm$0.35} & 25.68 {\scriptsize$\pm$0.19} & 58.46 {\scriptsize$\pm$0.15} \\
								Manifold Mixup & 87.04 {\scriptsize$\pm$0.34} & 78.90 {\scriptsize$\pm$0.33} & 42.16 {\scriptsize$\pm$0.30} & 26.01 {\scriptsize$\pm$0.16} & 58.53 {\scriptsize$\pm$0.15} \\
								\midrule
								IN & 86.67 {\scriptsize$\pm$0.20} & 76.17 {\scriptsize$\pm$0.24} & 47.25 {\scriptsize$\pm$0.21} & 24.79 {\scriptsize$\pm$0.15} & 58.72  {\scriptsize$\pm$0.10}\\
								IN + Inp. PTB. & 86.51 {\scriptsize$\pm$0.22} & 77.91 {\scriptsize$\pm$0.24} & 46.50 {\scriptsize$\pm$0.20} & 24.75 {\scriptsize$\pm$0.14} & 58.92 {\scriptsize$\pm$0.11}\\
								IN + Inp. Adv. & 88.04 {\scriptsize$\pm$0.21} & 79.04 {\scriptsize$\pm$0.24} & 44.76 {\scriptsize$\pm$0.21} & 25.92 {\scriptsize$\pm$0.15} & 59.23 {\scriptsize$\pm$0.12}\\
								IN + Rep. PTB. & 87.76 {\scriptsize$\pm$0.21} & 77.70 {\scriptsize$\pm$0.25} & 47.92 {\scriptsize$\pm$0.23} & 25.25 {\scriptsize$\pm$0.18} & 59.66 {\scriptsize$\pm$0.10}\\
								\midrule
								Meta BIN & 86.66 {\scriptsize$\pm$0.19} & 78.52 {\scriptsize$\pm$0.28} & 46.06 {\scriptsize$\pm$0.31} & 25.28 {\scriptsize$\pm$0.16} & 59.13 {\scriptsize$\pm$0.12} \\
								Fix. BN + IN & 86.66 {\scriptsize$\pm$0.21} & 77.79 {\scriptsize$\pm$0.27} & 48.67 {\scriptsize$\pm$0.29} & 25.26 {\scriptsize$\pm$0.17} & 59.59 {\scriptsize$\pm$0.13} \\
								Sep. BN + IN & 88.51 {\scriptsize$\pm$0.20} & 79.38 {\scriptsize$\pm$0.19} & 44.85 {\scriptsize$\pm$0.21} & 26.25 {\scriptsize$\pm$0.18} & 59.66 {\scriptsize$\pm$0.14} \\
								\midrule
								Ours & \textbf{89.35} {\scriptsize$\pm$0.17} & \textbf{79.40} {\scriptsize $\pm$0.27} & \textbf{50.75} {\scriptsize $\pm$0.30} & \textbf{26.57} {\scriptsize $\pm$0.16} & \textbf{61.52} {\scriptsize $\pm$0.12} \\
								\bottomrule
						\end{tabular}}
					\end{center}\vspace{-0.45cm}
				\end{table}
				
				We ablate each design in Tab.~\ref{tab:ablation} with the prototype classifier for fairness (i.e., fixing the backbone network, using Euclidean distance without domain calibration). 
				We can see:
				
				(1) \textbf{SAM on RSLL is better than PSLL}. We implement the sharpness-aware minimization (SAM~\cite{foret2021sharpnessaware}) on both the RSLL and PSLL, and see the performance on RSLL is higher, because RSLL handles domain shifts directly.
				
				(2) \textbf{IN can partly account for our improvements}. By simply replacing BN with IN, the performance increases, since IN leads to a flatter minimum in the representation-space loss landscapes (verified in Fig.~\ref{fig:plot_input_noise_train} and \ref{tab:mini-c}). But this is still much lower than ours.
				
				(3) \textbf{Our perturbation direction is effective}.
				Since the flattening can be viewed to perturb representations, we compare it with some perturbation methods. We first follow Fig.~\ref{fig:plot_input_noise_train} to add low-frequency noise for domain shifting on the input image (IN + Inp. PTB) and representations (IN + Rep. PTB). The result is higher than that of IN but lower than ours, verifying the effectiveness of low-frequency noise and experiments in Fig.~\ref{fig:plot_input_noise_train}. 
				An alternative is to perturb in the gradient direction (adversarial attack~\cite{chakraborty2018adversarial}), therefore we tried this option as IN + Inp Adv, whose performance is similar, due to the attacked semantics in samples
				
				(4) \textbf{Learnable mixing ratio is not effective enough}.
				Some works also combine the output of BN and IN outputs, generating a learnable mixing ratio. Meta BIN~\cite{choi2021meta} is chosen as a representative but its performance is also limited. This is because the learnable ratio simplifies the representation learning, while FLoR increases the difficulty as it additionally constrains the high-loss region between two minima.
				
				(5) \textbf{Random ratio is better}. We test with two setups: Fix. BN + IN means to manually specify a pre-defined $\delta$ for both the base-class training and novel-class evaluation, and Sep. BN + IN refers to the model in Fig.~\ref{fig:plot_inW_before+after}. We can see such simple designs can lead to better results than Meta BIN which utilizes a learnable weight. 
				However, since it can only cover one intermediate point while ours can randomly cover all points, our performance is better.
				
				\vspace{-0.1cm}
				\subsubsection{Verification of easier finetuning}
				\vspace{-0.1cm}
				
\begin{figure}[t]
					\centering
					\includegraphics[width=1.0\columnwidth]{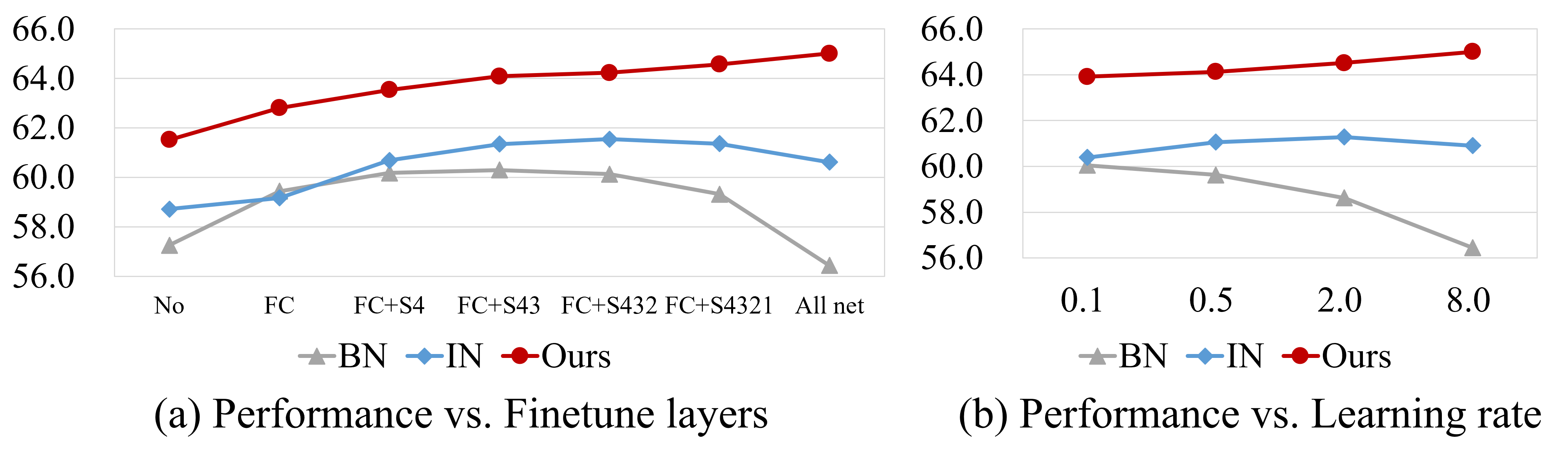}\vspace{-0.4cm}
					\caption{To verify the easy finetuning, we evaluate the finetuning with different layers and learning rates. FC: fully-connected layer, S: ResNet Stage, No: Prototype classifier without finetuning. Our model can directly finetune all layers with a large learning rate, indicating the easiness of few-shot finetuning.}\vspace{-0.4cm}
					\label{fig:finetune}
				\end{figure}

				We report the finetuning performance in Fig.~\ref{fig:finetune}. Typical finetuning methods~\cite{guobroader,zou2022margin} refer to finetuning the backbone with a small learning rate and fixing shallow layers due to the overfitting caused by scarce training data. We first test this claim by finetuning the BN baseline model (gray curve in Fig.~\ref{fig:finetune}). We can see this claim holds for this BN model. The best performance is achieved by fixing the first 3 stages and setting the learning rate to 0.1 times of the classifier, making the finetuning sensitive to hyper-parameter choices. 
				In contrast, the IN model shows more layers can be finetuned with a larger learning rate. For ours, we can directly apply all layers into the finetuning, with a large learning rate at 8 times that in the classifier. This means our model is easier to finetune under the few-shot scenarios and is not sensitive to the hyper-parameter choices, which is the consequence of the flat minimum as validated in Fig.~\ref{fig:plot_input_noise_train} and \ref{fig:plot_inW_before+after}.
				
				
				
				
				
				\vspace{-0.2cm}
				\section{Related Work}
				\vspace{-0.1cm}
				
				\subsection{Cross domain few-shot learning (CDFSL)}
				\vspace{-0.1cm}
				
				CDFSL aims to effectively learn from target domains with only a few training samples~\cite{DBLP:journals/corr/abs-1904-04232,guobroader}. Compared with the ordinary few-shot learning task, CDFSL needs to handle a larger domain gap, which makes it more difficult to transfer and finetune. Current works can be grouped into transferring-based~\cite{Zhou_2023_CVPR,wang2021crossdomain,hu2022adversarial} and finetuning-based works~\cite{li2022ranking,guobroader}. 
				Compared with them, our method focuses on source-domain training, but we can benefit the knowledge transfer and target-domain finetuning at the same time.
				
				Some works~\cite{tseng2020cross,Zhou_2023_CVPR} show that the BN module is vulnerable to domain shifts. We also follow this conclusion, but we are the first to study the loss landscape around BN representations.
				Mixup~\cite{zhang2018mixup,verma2019manifold} is another line of learning transferable models.
				Compared with them, our method only shares a similar formulation in interpolation but differs in (1) we do not need multiple samples for mixing, (2) we aim to flatten the loss landscape between two normalized representations, and (3) our results are much higher (Tab.~\ref{tab:ablation}). 
				
				\vspace{-0.1cm}
				\subsection{Generalization based on loss landscapes}
				\vspace{-0.1cm}
				
				Current works show the relationship between loss landscapes and the generalization~\cite{foret2021sharpnessaware}, where a widely adopted conclusion is that a flat minimum in loss landscapes refers to a more generalizable model. Most of these works analyze the loss landscapes from the \textbf{parameter space}~\cite{keskar2017largebatch} to handle the problem such as domain generalization~\cite{zhang2023gradient} and catastrophic forgetting~\cite{shi2021overcoming}. Compared with them, we are the first to introduce the analysis from the aspect of \textbf{representation-space} loss landscape into the CDFSL task. 

				\vspace{-0.1cm}
				\section{Conclusion}
				\vspace{-0.1cm}
				
				We analyze model generalization from representation-space loss landscapes for CDFSL, and propose to flatten the long-range loss landscapes by randomly interpolating the differently normalized representations. Experiments on 8 datasets validate the effectiveness and rationale.
				
				\vspace{-0.1cm}
				\section*{Acknowledgments}
				\vspace{-0.1cm}
				
				This work is supported by National Natural Science Foundation of China under grants 62206102, 62376103, 62302184, and Science and Technology Support Program of Hubei Province under grant 2022BAA046.

				{
					\small
					\bibliographystyle{ieeenat_fullname}
					\bibliography{main}
				}
				

\end{document}